\documentclass{article} 
\usepackage{iclr2025,times}

\usepackage{amsmath,amsfonts,bm}

\def\eqref#1{equation~\ref{#1}}

\def\1{\bm{1}}

\DeclareMathAlphabet{\mathsfit}{\encodingdefault}{\sfdefault}{m}{sl}
\SetMathAlphabet{\mathsfit}{bold}{\encodingdefault}{\sfdefault}{bx}{n}

\usepackage{mathtools}
\usepackage{hyperref}
\usepackage{url}

\usepackage[english]{babel}
\addto\captionsenglish{%
}
\addto\extrasenglish{%
}

\title{Last Layer Empirical Bayes}

\author{Valentin Villecroze\\
      Layer 6 AI\\
      \texttt{valentin.v@layer6.ai}
      \And
      Yixin Wang\\
      University of Michigan\\
      \texttt{yixinw@umich.edu}
      \And
      Gabriel Loaiza-Ganem\\
      Layer 6 AI\\
      \texttt{gabriel@layer6.ai}
      }

\iclrfinalcopy 
\begin{document}

\maketitle

\begin{abstract}
The task of quantifying the inherent uncertainty associated with neural network predictions is a key challenge in artificial intelligence. Bayesian neural networks (BNNs) and deep ensembles are among the most prominent approaches to tackle this task. Both approaches produce predictions by computing an expectation of neural network outputs over some distribution on the corresponding weights; this distribution is given by the posterior in the case of BNNs, and by a mixture of point masses for ensembles. Inspired by recent work showing that the distribution used by ensembles can be understood as a posterior corresponding to a learned data-dependent prior, we propose last layer empirical Bayes (LLEB). LLEB instantiates a learnable prior as a normalizing flow, which is then trained to maximize the evidence lower bound; to retain tractability we use the flow only on the last layer. We show why LLEB is well motivated, and how it interpolates between standard BNNs and ensembles in terms of the strength of the prior that they use. LLEB performs on par with existing approaches, highlighting that empirical Bayes is a promising direction for future research in uncertainty quantification.
\end{abstract}

\section{Introduction}\label{sec:intro}
Uncertainty quantification (UQ) is a crucial task in scientific and safety-critical settings \citep{esteva2017dermatologist, bojarski2016end, litjens2017survey, psaros2023uncertainty}, and every improvement in UQ within deep learning is a step towards a broader adoption of AI. The two most popular approaches for UQ are Bayesian neural networks \citep[BNNs;][]{welling2011bayesian, graves2011practical, hernandez2015probabilistic, blundell2015weight, gal2016dropout, ritter2018scalable} and deep ensembles \citep{lakshminarayanan2017simple}. Both BNNs and ensembles produce a distribution $q^*$ over neural network weights $\theta$; $q^*$ is the Bayesian posterior (or an approximation thereof) for BNNs, and a mixture of point masses obtained by independent training runs for ensembles. Computing the expectation of network outputs over $\theta \sim q^*$ produces predictions, and the corresponding variability of the outputs over $\theta$ can be used to quantify uncertainty.

Ensembles typically outperform BNNs at UQ, but they do so at increased computational cost \citep{Abdar2021Survey}. In recent work, \citet{loaiza-ganem2025deep} pointed out that the distribution $q^*$ used by ensembles can actually be interpreted as a Bayesian posterior corresponding to a learned data-dependent prior. In this sense ensembles are BNNs, except the prior is not fixed beforehand as in standard BNNs. \citet{loaiza-ganem2025deep} also argue that using priors which concentrate their mass around the set of maximum-likelihood weights is likely beneficial for UQ and a potential reason behind the good performance of ensembles. Inspired by this connection, we propose last layer empirical Bayes (LLEB) as an intermediate between standard BNNs and ensembles in terms of the strength of the used prior: LLEB is a BNN where the prior is instantiated as a normalizing flow \citep[NFs;][]{dinh2015nice, rezende2015variational, durkan2019neural} and learned by maximizing the standard evidence lower bound (ELBO) from variational inference \citep[VI;][]{wainwright2008graphical, kingma2014auto, rezende2014stochastic, blei2017variational}. One key motivation behind LLEB is that by learning the prior through a NF, $q^*$ can place most of its mass around ``good values of $\theta$'' while retaining diversity in $\theta$, thus hopefully achieving comparable performance to ensembles without the need to train various models. To maintain tractability, we follow the recent trend in BNNs of being Bayesian only over a subset of parameters such as those in the last layer \citep{lazaro2010marginalized, kristiadi2020being, watson2020neural, watson2021latent, harrison2024variational, yang2024bayesian}.

Empirically, we find that LLEB performs on par but not significantly nor consistently better than existing UQ approaches of similar computational cost. Our results highlight the promise of empirical Bayes for UQ, and we hope that future work will be able to leverage the ideas behind LLEB to outperform existing UQ methods.

\section{Background and Related Work}
\paragraph{Setup} Throughout this work we will consider a classification setup, where we have access to a dataset $\mathcal{D}=\{(x_i, y_i)\}_{i}$ of feature-label pairs $(x_i, y_i)$. Here, the likelihood $p(\mathcal{D} \mid \theta)$ is given by $p(\mathcal{D} \mid \theta) = \prod_{i} p(y_i \mid x_i, \theta)$, where $p(y_i \mid x_i, \theta)$ is the probability assigned by the neural network parameterized by $\theta \in \Theta$ to the label $y_i$ when given the input $x_i$. We will assume that the likelihood function achieves its maximum, and will denote the set of maximizers as $\Theta^* \subset \Theta$.

\paragraph{Bayesian neural networks and variational inference} BNNs begin by specifying a prior $\pi$, often as a Gaussian with diagonal covariance. The main object of interest in BNNs is then the corresponding posterior distribution, which is given by $\pi(\theta \mid \mathcal{D}) \propto \pi(\theta)p(\mathcal{D} \mid \theta)$. Unfortunately, computing $\pi(\, \cdot \mid \mathcal{D})$ and sampling from it are intractable. Various lines of research, which we will shortly summarize, attempt to circumvent this problem by providing a distribution $q^*$ whose goal is to approximate the posterior, i.e.\ $q^* \approx \pi(\, \cdot \mid \mathcal{D})$. Once $q^*$ has been obtained, the epistemic uncertainty \citep{hullermeier2021aleatoric} associated with predicting the label of a query point $x_{n+1}$ can be quantified through the variability of $p(\, \cdot \mid x_{n+1}, \theta)$ over $\theta \sim q^*$, and predictions can be made through the predictive distribution,
\begin{equation}\label{eq:predictive}
    p(\, \cdot \mid x_{n+1}) \coloneqq \mathbb{E}_{\theta \sim q^*}\left[p(\, \cdot \mid x_{n+1}, \theta)\right].
\end{equation}
One class of methods uses a Laplace approximation, i.e.\ a second-order Taylor expansion of \hbox{$\log \pi(\, \cdot \mid \mathcal{D})$}, to obtain $q^*$ \citep{ritter2018scalable, kristiadi2020being, daxberger2021laplace, yang2024bayesian}; this results in $q^*$ being a Gaussian approximation of the posterior. Another class of methods uses Markov chain Monte Carlo to approximately sample from $\pi(\, \cdot \mid \mathcal{D})$ \citep{welling2011bayesian, chen2014stochastic, zhang2020cyclical}, here the distribution of the chain corresponds to $q^*$. \citet{gal2016dropout} obtain $q^*$ by using dropout \citep{srivastava2014dropout}.

A final relevant class of procedures to obtain $q^*$ do so through VI \citep{graves2011practical, blundell2015weight, louizos2016structured, louizos2017multiplicative, wu2019deterministic, osawa2019practical, harrison2024variational}, i.e.\ by maximizing the ELBO,
\begin{equation}\label{eq:elbo}
    \text{ELBO}(q, \pi) \coloneqq \mathbb{E}_{\theta \sim q}\left[\log p(\mathcal{D} \mid \theta)\right] - \mathbb{KL}\left(q \Vert \pi \right),
\end{equation}
over $q \in \mathcal{Q}$ for some family of distributions $\mathcal{Q}$. When $\mathcal{Q}$ is flexible enough in the sense that it contains the true posterior, this maximization is well known to yield $q^* = \pi(\, \cdot \mid \mathcal{D})$. However, most VI-based BNN methods use simple choices of $\mathcal{Q}$ (e.g.\ Gaussians) due to tractability. Many methods use NFs\footnote{Recall that NFs define a density $q$ as the density of $f(Z)$, where $f$ is an invertible neural network and $Z$ has a simple distribution such as an isotropic Gaussian.} to instantiate $\mathcal{Q}$ \citep{rezende2015variational, kingma2016improved}, resulting in increased flexibility while keeping the KL term in the ELBO tractable. However, these methods apply VI in the context of variational autoencoders \citep{kingma2014auto, rezende2014stochastic} and not to BNNs because the size of NFs cannot be scaled to the number of parameters in a neural network. Indeed, the large number of parameters in neural networks results in BNNs having to deal with extremely high-dimensional distributions; some works have sought to circumvent this bottleneck by being Bayesian only over the last layer of the network \citep{kristiadi2020being, harrison2024variational}.

\paragraph{Deep ensembles and their connection to Bayesian neural networks} Like BNNs, ensembles find a distribution $q^*$ which is also used to quantify uncertainty, and to produce predictions through \autoref{eq:predictive}. Ensembles train $M$ separate models through maximum-likelihood, i.e.\ maximizing $\log p(\mathcal{D} \mid \theta)$, to obtain $\theta_m^* \in \Theta^*$ for $m=1,\dots,M$; all these values are different due to the randomness of stochastic optimization -- the resulting $q^*$ is then given by $q^*(\theta) = \tfrac{1}{M}\sum_{m} \delta_{\theta_m^*}(\theta)$, where $\delta_{\theta_m^*}$ denotes a point mass at $\theta_m^*$. Although ensembles are not typically thought of as Bayesian, \citet{loaiza-ganem2025deep} recently argued they can be understood as performing empirical Bayes, i.e.\ learning the prior $\pi$ from data. More specifically, assuming enough capacity, the ELBO in \autoref{eq:elbo} is maximized over both $q$ and $\pi$ (rather than just $q$) by $(q^*, \pi^*)$ if and only if $q^*$ assigns probability $1$ to $\Theta^*$ and $q^*=\pi^*$; in this case, the prior $\pi^*$, its corresponding posterior \hbox{$\pi^*(\, \cdot \mid \mathcal{D})$}, and $q^*$ are all equal to each other (see \autoref{app:extra_background} for more details). The particular $q^*$ used by ensembles assigns probability $1$ to $\Theta^*$, and thus it follows that it can be interpreted as both a learned prior and its corresponding posterior. In short, the main difference between BNNs and ensembles is that BNNs use weak (e.g.\ Gaussian), fixed priors (or with at most the variance being learnable), whereas ensembles use strong and implicitly learned data-dependent priors.

\section{Last Layer Empirical Bayes}\label{sec:method}
There are three main motivations behind our work. First, deep ensembles tend to outperform BNNs at UQ \citep{Abdar2021Survey}, and the empirical Bayes view of ensembles thus suggests that using stronger, data-dependent priors is preferable to using weak, fixed ones. Consequently, we aim to explore explicitly learning the prior. Second, although the empirical Bayes view of ensembles suggests that very strong priors are better than very weak ones, it does not guarantee that stronger is always better. In particular, the prior $q^*$ used by ensembles is extremely strong, and part of our motivation is to use a slightly weaker learned prior which is still strong enough to concentrate mass around $\Theta^*$. Third, ensembles are computationally expensive as they require training $M$ models. Our final motivator is that by explicitly learning $q^*$ once we can avoid training $M$ models. We hope that a model satisfying these motivations might perform similarly to ensembles while being cheaper to train.

With these motivations in mind, we first consider simply maximizing $\mathbb{E}_{\theta \sim q}[\log p(\mathcal{D} \mid \theta)]$ over \hbox{$q \in \mathcal{Q}$}; this will produce the same optimal $q^*$ as maximizing $\text{ELBO}(q, \pi)$ over $q$ and $\pi$ under a flexible enough $\pi$. Furthermore, if $\mathcal{Q}$ is flexible enough, the resulting $q^*$ will assign probability $1$ to $\Theta^*$, and could thus be interpreted as both a prior and its corresponding posterior, just like in ensembles. Our goal here is then to choose a $\mathcal{Q}$ which $(i)$ is flexible enough for $q^*$ to concentrate mass around $\Theta^*$ while not collapsing onto a point mass (as this would just recover a maximum-likelihood solution), and $(ii)$ results in $q^*$ being more diverse than the mixture of point masses used by ensembles. Specifying $\mathcal{Q}$ as NFs is then very natural since NFs are very flexible, yet their invertibility acts as an implicit regularizer which prevents collapse onto a point mass and promotes some diversity. In summary, we would ideally like to instantiate $q_\eta$ as a NF parameterized by $\eta$ and maximize $\mathbb{E}_{\theta \sim q_\eta}[\log p(\mathcal{D} \mid \theta)]$ over $\eta$ to then treat the resulting $q_{\eta^*}$ as we would $q^*$ in BNNs or ensembles.

Using $q_\eta$ as described above would satisfy our first two motivations but would still result in a highly intractable procedure despite not requiring to train $M$ models. The root cause of this intractability is the high dimensionality of the NF, and we thus propose to quantify uncertainty only over a subset of parameters. More precisely, let $\theta = (\theta_{QU}, \theta_{NU})$, where $\theta_{QU}$ and $\theta_{NU}$ are the parameters over which we do and do not quantify uncertainty, respectively. As a first attempt to address the tractability issues, we then instantiate $q_\eta$ as a NF on $\theta_{QU}$ and maximize $\mathbb{E}_{\theta_{QU}\sim q_\eta}[\log p(\mathcal{D} \mid \theta_{QU}, \theta_{NU})]$ over $\theta_{NU}$ and $\eta$. We found this end-to-end objective performed well with small classifiers, but that it resulted in unstable and slow optimization when using larger classifiers. As a way to circumvent this issue, we train our model in two steps: we first perform maximum likelihood by maximizing $\log p(\mathcal{D} \mid \theta_{QU}, \theta_{NU})$ to obtain $\theta_{QU}^*$ and $\theta_{NU}^*$, and we then discard $\theta_{QU}^*$ and maximize $\mathbb{E}_{\theta_{QU}\sim q_\eta}[\log p(\mathcal{D} \mid \theta_{QU}, \theta_{NU}^*)]$ over $\eta$; we found this strategy to be faster and much more stable for larger classifiers. Note that here $q^*$ is now formally given by $q^*(\theta_{QU}, \theta_{NU}) = \delta_{\theta_{NU}^*}(\theta_{NU})q_{\eta^*}(\theta_{QU})$.

In practice we chose to set $\theta_{UQ}$ as the weights of the last layer of the classifier. The resulting methods (both end-to-end and two-step training), which we call \emph{last layer empirical Bayes}, satisfy all our motivations: $q^*$ is flexible and explicitly learned, the invertibility of the flow prevents collapse onto point masses and encourages some diversity, and since the NF is relatively low-dimensional, training it does not incur significant computational overhead as compared to just maximizing the likelihood. We highlight that LLEB is certainly not the only way to satisfy our starting motivations; we include in \autoref{app:alternatives} various alternatives to LLEB which we considered but found to empirically underperform LLEB.

\section{Experiments}
\paragraph{Setup} We conduct experiments on two pairs of datasets, MNIST \citep{lecun1998mnist} \& Fashion-MNIST \citep{xiao2017fashion}, and CIFAR-10 \citep{krizhevsky2009learning} \& SVHN \citep{netzer2011reading}. For each pair, we use one dataset for the train and test sets, and the other as an out-of-distribution (OOD) set. For each pair we fix an architecture and compare LLEB against: the default network, last layer Laplace approximation \citep[LLL;][]{daxberger2021laplace}, and Monte Carlo dropout \citep[MCD;][]{gal2016dropout}; all these baselines have comparable computational costs to LLEB. We also compare ensembles of LLEB models against standard ensembles \citep{lakshminarayanan2017simple} and ensembled versions of all the aforementioned baselines; once again all these comparisons are fair from a perspective of computational cost. See \autoref{app:details} for more information on the implementation details; our code is available at \url{https://github.com/layer6ai-labs/last_layer_empirical_bayes}.

\paragraph{Metrics} We report the accuracy (Acc.) and the expected calibration error (ECE) over the test set; the latter measures how well models quantify aleatoric uncertainty \citep{hullermeier2021aleatoric}. For every test and OOD point $x$ we also compute $\sum_y \text{var}_{\theta \sim q^*}[p(y \mid x, \theta)]$ to quantify epistemic uncertainty; to evaluate how well models quantify epistemic uncertainty, we compute the area under the receiver operating characteristic curve (AUC) obtained when using this metric to classify between in- and out-of-distribution data (with large values corresponding to OOD). These metrics, along with standard errors across 5 random seeds, are shown in \autoref{tab:mnist_tab} and \autoref{tab:cifar_tab}.

\begin{table}[t]
    \centering
    \caption{\small{Results on MNIST \& Fashion-MNIST. The top and bottom parts show single and ensembled ($M=5$) models, respectively. For each metric, the best results within models of comparable computational cost are bolded (only best mean values are bolded).}}
    \resizebox{0.87\textwidth}{!}{
    \begin{tabular}{l|ccc|ccc}
        & \multicolumn{3}{c|}{Train/Test: MNIST, OOD: Fashion-MNIST} & \multicolumn{3}{c}{Train/Test: Fashion-MNIST, OOD: MNIST} \\
        \hline
        Method & Acc. ($\uparrow$) & ECE ($\downarrow$) & AUC ($\uparrow$) & Acc. ($\uparrow$) & ECE ($\downarrow$) & AUC ($\uparrow$) \\
         \hline
Default & $98.02 \pm 0.05$ & $\mathbf{0.00 \pm 0.00}$ & - & $88.02 \pm 0.10$ & $\mathbf{0.01 \pm 0.00}$ & - \\
LLL & $98.02 \pm 0.05$ & $0.75 \pm 0.00$ &  $\mathbf{0.96 \pm 0.00}$ & $88.02 \pm 0.10$ & $0.66 \pm 0.00$ & $\mathbf{0.82 \pm 0.01}$ \\
MCD & $\mathbf{98.50 \pm 0.05}$ & $0.01 \pm 0.00$ & $0.91 \pm 0.00$ & $\mathbf{88.47 \pm 0.05}$ & $0.02 \pm 0.00$ &  $0.75 \pm 0.03$ \\
LLEB (ours) & $97.74 \pm 0.24$ & $\mathbf{0.00 \pm 0.00}$ & $0.95 \pm 0.01$ & $87.83 \pm 0.37$ & $\mathbf{0.01 \pm 0.00}$ & $0.72 \pm 0.03$ \\
\hline
Default ($M=5$) & $98.26 \pm 0.02$ & $\mathbf{0.01 \pm 0.00}$ & $\mathbf{0.97 \pm 0.00}$ & $88.71 \pm 0.09$ & $\mathbf{0.02 \pm 0.00}$ &  $0.84 \pm 0.01$ \\
LLL ($M=5$) & $98.26 \pm 0.02$ & $0.76 \pm 0.00$ & $0.96 \pm 0.00$ & $88.71 \pm 0.09$ & $0.67 \pm 0.00$ & $0.87 \pm 0.01$ \\
MCD ($M=5$) & $\mathbf{98.69} \pm 0.02$ & $0.02 \pm 0.00$ & $0.95 \pm 0.00$ & $89.34 \pm 0.08$ & $0.04 \pm 0.00$ & $\mathbf{0.89 \pm 0.00}$ \\
LLEB ($M=5$, ours) & $98.30 \pm 0.08$ & $\mathbf{0.01 \pm 0.00}$ &  $\mathbf{0.97 \pm 0.00}$ & $\mathbf{89.44 \pm 0.16}$ & $0.03 \pm 0.00$ & $\mathbf{0.89 \pm 0.01}$ \\
        \hline
    \end{tabular}
    }
    \label{tab:mnist_tab}
\end{table}

\begin{table}[t]
    \centering
    \caption{\small{Results on CIFAR-10 \& SVHN, metrics and methods are identical to those in \autoref{tab:mnist_tab}.}}
    \resizebox{0.87\textwidth}{!}{
    \begin{tabular}{l|ccc|ccc}
        & \multicolumn{3}{c|}{Train/Test: CIFAR-10, OOD: SVHN} & \multicolumn{3}{c}{Train/Test: SVHN, OOD: CIFAR-10} \\
        \hline
        Method & Acc. ($\uparrow$) & ECE ($\downarrow$) & AUC ($\uparrow$) & Acc. ($\uparrow$) & ECE ($\downarrow$) &  AUC ($\uparrow$) \\
         \hline
Default & $92.82 \pm 0.09$ & $\mathbf{0.05 \pm 0.00}$ &  - & $\mathbf{95.26 \pm 0.03}$ & $0.03 \pm 0.00$ & - \\
LLL & $92.82 \pm 0.09$ & $0.70 \pm 0.00$ & $\mathbf{0.94 \pm 0.01}$ & $\mathbf{95.26 \pm 0.03}$ & $0.73 \pm 0.00$ & $\mathbf{0.92 \pm 0.00}$ \\
MCD & $92.29 \pm 0.09$ & $0.10 \pm 0.01$ & $0.89 \pm 0.02$ & $95.11 \pm 0.05$ & $0.09 \pm 0.01$ & $0.89 \pm 0.00$  \\
LLEB (ours) & $\mathbf{92.85 \pm 0.09}$ & $0.06 \pm 0.00$ & $\mathbf{0.94 \pm 0.01}$ & $95.23 \pm 0.03$ & $\mathbf{0.02 \pm 0.01}$ & $0.86 \pm 0.01$ \\
\hline
Default ($M=5$) & $\mathbf{94.82 \pm 0.01}$ & $\mathbf{0.01 \pm 0.00}$ & $0.91 \pm 0.01$ & $\mathbf{96.55 \pm 0.03}$ & $\mathbf{0.01 \pm 0.00}$ & $0.97 \pm 0.00$ \\
LLL ($M=5$) & $\mathbf{94.82 \pm 0.01}$ & $0.73 \pm 0.00$ & $0.90 \pm 0.01$ & $\mathbf{96.55 \pm 0.03}$ & $0.74 \pm 0.00$ & $0.97 \pm 0.00$ \\
MCD ($M=5$) & $94.72 \pm 0.04$ & $0.12 \pm 0.00$ & $0.93 \pm 0.01$ & $96.54 \pm 0.02$ & $0.11 \pm 0.00$ & $\mathbf{0.98 \pm 0.00}$ \\
LLEB ($M=5$, ours) & $94.78 \pm 0.01$ & $\mathbf{0.01 \pm 0.00}$ & $\mathbf{0.95 \pm 0.01}$ & $96.52 \pm 0.03$ & $\mathbf{0.01 \pm 0.00}$ & $\mathbf{0.98 \pm 0.00}$ \\
        \hline
    \end{tabular}
    }
    \label{tab:cifar_tab}
\end{table}

\paragraph{Results} LLEB underperforms ensembles despite its ambitious motivation being to achieve similar results, nevertheless we stress that this comparison favours ensembles since they are much more costly to train. 
Although LLEB outperforms baselines of comparable computational cost in a few tasks and metrics, it does not do so consistently; this holds true both for single models and when ensembling. We see LLEB performing on par with the best existing baselines as highlighting the promise in its empirical Bayes motivation, yet we also believe that LLEB not outperforming these baselines is likely a consequence of the choices we made for tractability.

\section{Conclusion}
In this work we argued that maximizing $\mathbb{E}_{\theta \sim q}[\log p(\mathcal{D} \mid \theta)]$ over $q$ (with potential regularization) is backed by empirical Bayes as a sensible approach towards UQ. We further proposed LLEB as a way to approximately maximize this objective while retaining tractability. Although LLEB has decent performance, it does not significantly outperform other UQ methods; we hypothesize that this is due to the concessions we made in LLEB for tractability. We hope that future research will manage to improve upon LLEB by better leveraging empirical Bayes to learn $q^*$ in a way that remains tractable and outperforms existing UQ approaches which use simple priors.


\subsubsection*{Acknowledgments}
We thank Brendan Ross for insightful discussions and for having provided feedback on our manuscript.

\bibliography{iclr2025}

\begin{thebibliography}{44}
\providecommand{\natexlab}[1]{#1}
\providecommand{\url}[1]{\texttt{#1}}
\expandafter\ifx\csname urlstyle\endcsname\relax
  \providecommand{\doi}[1]{doi: #1}\else
  \providecommand{\doi}{doi: \begingroup \urlstyle{rm}\Url}\fi

\bibitem[Abdar et~al.(2021)Abdar, Pourpanah, Hussain, Rezazadegan, Liu, Ghavamzadeh, Fieguth, Cao, Khosravi, Acharya, Makarenkov, and Nahavandi]{Abdar2021Survey}
Moloud Abdar, Farhad Pourpanah, Sadiq Hussain, Dana Rezazadegan, Li~Liu, Mohammad Ghavamzadeh, Paul~W. Fieguth, Xiaochun Cao, Abbas Khosravi, U.~Rajendra Acharya, Vladimir Makarenkov, and Saeid Nahavandi.
\newblock A review of uncertainty quantification in deep learning: Techniques, applications and challenges.
\newblock \emph{Information Fusion}, 76:\penalty0 243--297, 2021.

\bibitem[Blei et~al.(2017)Blei, Kucukelbir, and McAuliffe]{blei2017variational}
David~M Blei, Alp Kucukelbir, and Jon~D McAuliffe.
\newblock Variational inference: A review for statisticians.
\newblock \emph{Journal of the American statistical Association}, 112\penalty0 (518):\penalty0 859--877, 2017.

\bibitem[Blundell et~al.(2015)Blundell, Cornebise, Kavukcuoglu, and Wierstra]{blundell2015weight}
Charles Blundell, Julien Cornebise, Koray Kavukcuoglu, and Daan Wierstra.
\newblock Weight uncertainty in neural network.
\newblock In \emph{International Conference on Machine Learning}, 2015.

\bibitem[Bojarski et~al.(2016)Bojarski, Del~Testa, Dworakowski, Firner, Flepp, Goyal, Jackel, Monfort, Muller, Zhang, Zhang, Zhao, and Zieba]{bojarski2016end}
Mariusz Bojarski, Davide Del~Testa, Daniel Dworakowski, Bernhard Firner, Beat Flepp, Prasoon Goyal, Lawrence~D Jackel, Mathew Monfort, Urs Muller, Jiakai Zhang, Xin Zhang, Jake Zhao, and Karol Zieba.
\newblock End to end learning for self-driving cars.
\newblock \emph{arXiv:1604.07316}, 2016.

\bibitem[Chen et~al.(2014)Chen, Fox, and Guestrin]{chen2014stochastic}
Tianqi Chen, Emily Fox, and Carlos Guestrin.
\newblock Stochastic gradient hamiltonian monte carlo.
\newblock In \emph{International Conference on Machine Learning}, 2014.

\bibitem[Daxberger et~al.(2021)Daxberger, Kristiadi, Immer, Eschenhagen, Bauer, and Hennig]{daxberger2021laplace}
Erik Daxberger, Agustinus Kristiadi, Alexander Immer, Runa Eschenhagen, Matthias Bauer, and Philipp Hennig.
\newblock Laplace redux-effortless {B}ayesian deep learning.
\newblock In \emph{Advances in Neural Information Processing Systems}, 2021.

\bibitem[Dinh et~al.(2015)Dinh, Krueger, and Bengio]{dinh2015nice}
Laurent Dinh, David Krueger, and Yoshua Bengio.
\newblock {NICE}: Non-linear independent components estimation.
\newblock In \emph{ICLR Workshop Track}, 2015.

\bibitem[Durkan et~al.(2019)Durkan, Bekasov, Murray, and Papamakarios]{durkan2019neural}
Conor Durkan, Artur Bekasov, Iain Murray, and George Papamakarios.
\newblock Neural spline flows.
\newblock In \emph{Advances in Neural Information Processing Systems}, 2019.

\bibitem[Durkan et~al.(2020)Durkan, Bekasov, Murray, and Papamakarios]{nflows}
Conor Durkan, Artur Bekasov, Iain Murray, and George Papamakarios.
\newblock {nflows}: normalizing flows in {PyTorch}, November 2020.
\newblock URL \url{https://doi.org/10.5281/zenodo.4296287}.

\bibitem[Esteva et~al.(2017)Esteva, Kuprel, Novoa, Ko, Swetter, Blau, and Thrun]{esteva2017dermatologist}
Andre Esteva, Brett Kuprel, Roberto~A Novoa, Justin Ko, Susan~M Swetter, Helen~M Blau, and Sebastian Thrun.
\newblock Dermatologist-level classification of skin cancer with deep neural networks.
\newblock \emph{Nature}, 542\penalty0 (7639):\penalty0 115--118, 2017.

\bibitem[Gal \& Ghahramani(2016)Gal and Ghahramani]{gal2016dropout}
Yarin Gal and Zoubin Ghahramani.
\newblock Dropout as a {B}ayesian approximation: Representing model uncertainty in deep learning.
\newblock In \emph{International Conference on Machine Learning}, 2016.

\bibitem[Graves(2011)]{graves2011practical}
Alex Graves.
\newblock Practical variational inference for neural networks.
\newblock In \emph{Advances in Neural Information Processing Systems}, 2011.

\bibitem[Harrison et~al.(2024)Harrison, Willes, and Snoek]{harrison2024variational}
James Harrison, John Willes, and Jasper Snoek.
\newblock Variational {B}ayesian last layers.
\newblock In \emph{International Conference on Learning Representations}, 2024.

\bibitem[He et~al.(2016)He, Zhang, Ren, and Sun]{he2016deep}
Kaiming He, Xiangyu Zhang, Shaoqing Ren, and Jian Sun.
\newblock Deep residual learning for image recognition.
\newblock In \emph{Proceedings of the IEEE Conference on Computer Vision and Pattern Recognition}, 2016.

\bibitem[Hern{\'a}ndez-Lobato \& Adams(2015)Hern{\'a}ndez-Lobato and Adams]{hernandez2015probabilistic}
Jos{\'e}~Miguel Hern{\'a}ndez-Lobato and Ryan Adams.
\newblock Probabilistic backpropagation for scalable learning of {B}ayesian neural networks.
\newblock In \emph{International Conference on Machine Learning}, 2015.

\bibitem[H{\"u}llermeier \& Waegeman(2021)H{\"u}llermeier and Waegeman]{hullermeier2021aleatoric}
Eyke H{\"u}llermeier and Willem Waegeman.
\newblock Aleatoric and epistemic uncertainty in machine learning: An introduction to concepts and methods.
\newblock \emph{Machine Learning}, 110\penalty0 (3):\penalty0 457--506, 2021.

\bibitem[Kingma \& Ba(2015)Kingma and Ba]{kingma2015adam}
Diederik~P Kingma and Jimmy Ba.
\newblock Adam: A method for stochastic optimization.
\newblock In \emph{International Conference on Learning Representations}, 2015.

\bibitem[Kingma \& Welling(2014)Kingma and Welling]{kingma2014auto}
Diederik~P Kingma and Max Welling.
\newblock {Auto-encoding variational Bayes}.
\newblock In \emph{International Conference on Learning Representations}, 2014.

\bibitem[Kingma et~al.(2016)Kingma, Salimans, Jozefowicz, Chen, Sutskever, and Welling]{kingma2016improved}
Diederik~P Kingma, Tim Salimans, Rafal Jozefowicz, Xi~Chen, Ilya Sutskever, and Max Welling.
\newblock Improved variational inference with inverse autoregressive flow.
\newblock In \emph{Advances in Neural Information Processing Systems}, 2016.

\bibitem[Kristiadi et~al.(2020)Kristiadi, Hein, and Hennig]{kristiadi2020being}
Agustinus Kristiadi, Matthias Hein, and Philipp Hennig.
\newblock Being {B}ayesian, even just a bit, fixes overconfidence in relu networks.
\newblock In \emph{International Conference on Machine Learning}, 2020.

\bibitem[Krizhevsky \& Hinton(2009)Krizhevsky and Hinton]{krizhevsky2009learning}
Alex Krizhevsky and Geoffrey Hinton.
\newblock Learning multiple layers of features from tiny images.
\newblock 2009.

\bibitem[Lakshminarayanan et~al.(2017)Lakshminarayanan, Pritzel, and Blundell]{lakshminarayanan2017simple}
Balaji Lakshminarayanan, Alexander Pritzel, and Charles Blundell.
\newblock Simple and scalable predictive uncertainty estimation using deep ensembles.
\newblock In \emph{Advances in Neural Information Processing Systems}, 2017.

\bibitem[L{\'a}zaro-Gredilla \& Figueiras-Vidal(2010)L{\'a}zaro-Gredilla and Figueiras-Vidal]{lazaro2010marginalized}
Miguel L{\'a}zaro-Gredilla and An{\'\i}bal~R Figueiras-Vidal.
\newblock Marginalized neural network mixtures for large-scale regression.
\newblock \emph{IEEE transactions on neural networks}, 21\penalty0 (8):\penalty0 1345--1351, 2010.

\bibitem[LeCun et~al.(1998)LeCun, Bottou, Bengio, and Haffner]{lecun1998mnist}
Yann LeCun, L{\'e}on Bottou, Yoshua Bengio, and Patrick Haffner.
\newblock Gradient-based learning applied to document recognition.
\newblock \emph{Proceedings of the IEEE}, 86\penalty0 (11):\penalty0 2278--2324, 1998.

\bibitem[Litjens et~al.(2017)Litjens, Kooi, Bejnordi, Setio, Ciompi, Ghafoorian, Van Der~Laak, Van~Ginneken, and S{\'a}nchez]{litjens2017survey}
Geert Litjens, Thijs Kooi, Babak~Ehteshami Bejnordi, Arnaud Arindra~Adiyoso Setio, Francesco Ciompi, Mohsen Ghafoorian, Jeroen~Awm Van Der~Laak, Bram Van~Ginneken, and Clara~I S{\'a}nchez.
\newblock A survey on deep learning in medical image analysis.
\newblock \emph{Medical Image Analysis}, 42:\penalty0 60--88, 2017.

\bibitem[Loaiza-Ganem et~al.(2017)Loaiza-Ganem, Gao, and Cunningham]{loaiza2017maximum}
Gabriel Loaiza-Ganem, Yuanjun Gao, and John~P Cunningham.
\newblock Maximum entropy flow networks.
\newblock In \emph{International Conference on Learning Representations}, 2017.

\bibitem[Loaiza-Ganem et~al.(2025)Loaiza-Ganem, Villecroze, and Wang]{loaiza-ganem2025deep}
Gabriel Loaiza-Ganem, Valentin Villecroze, and Yixin Wang.
\newblock Deep ensembles secretly perform empirical bayes.
\newblock \emph{arXiv:2501.17917}, 2025.

\bibitem[Louizos \& Welling(2016)Louizos and Welling]{louizos2016structured}
Christos Louizos and Max Welling.
\newblock Structured and efficient variational deep learning with matrix {G}aussian posteriors.
\newblock In \emph{International Conference on Machine Learning}, 2016.

\bibitem[Louizos \& Welling(2017)Louizos and Welling]{louizos2017multiplicative}
Christos Louizos and Max Welling.
\newblock Multiplicative normalizing flows for variational {B}ayesian neural networks.
\newblock In \emph{International Conference on Machine Learning}, 2017.

\bibitem[Netzer et~al.(2011)Netzer, Wang, Coates, Bissacco, Wu, and Ng]{netzer2011reading}
Yuval Netzer, Tao Wang, Adam Coates, Alessandro Bissacco, Bo~Wu, and Andrew~Y Ng.
\newblock Reading digits in natural images with unsupervised feature learning.
\newblock In \emph{NIPS Workshop on Deep Learning and Unsupervised Feature Learning}, 2011.

\bibitem[Osawa et~al.(2019)Osawa, Swaroop, Khan, Jain, Eschenhagen, Turner, and Yokota]{osawa2019practical}
Kazuki Osawa, Siddharth Swaroop, Mohammad Emtiyaz~E Khan, Anirudh Jain, Runa Eschenhagen, Richard~E Turner, and Rio Yokota.
\newblock Practical deep learning with {B}ayesian principles.
\newblock In \emph{Advances in Neural Information Processing Systems}, 2019.

\bibitem[Psaros et~al.(2023)Psaros, Meng, Zou, Guo, and Karniadakis]{psaros2023uncertainty}
Apostolos~F Psaros, Xuhui Meng, Zongren Zou, Ling Guo, and George~Em Karniadakis.
\newblock Uncertainty quantification in scientific machine learning: Methods, metrics, and comparisons.
\newblock \emph{Journal of Computational Physics}, 477:\penalty0 111902, 2023.

\bibitem[Rezende \& Mohamed(2015)Rezende and Mohamed]{rezende2015variational}
Danilo~Jimenez Rezende and Shakir Mohamed.
\newblock Variational inference with normalizing flows.
\newblock In \emph{International Conference on Machine Learning}, 2015.

\bibitem[Rezende et~al.(2014)Rezende, Mohamed, and Wierstra]{rezende2014stochastic}
Danilo~Jimenez Rezende, Shakir Mohamed, and Daan Wierstra.
\newblock Stochastic backpropagation and approximate inference in deep generative models.
\newblock In \emph{International Conference on Machine Learning}, 2014.

\bibitem[Ritter et~al.(2018)Ritter, Botev, and Barber]{ritter2018scalable}
Hippolyt Ritter, Aleksandar Botev, and David Barber.
\newblock A scalable {L}aplace approximation for neural networks.
\newblock In \emph{International Conference on Learning Representations}, 2018.

\bibitem[Srivastava et~al.(2014)Srivastava, Hinton, Krizhevsky, Sutskever, and Salakhutdinov]{srivastava2014dropout}
Nitish Srivastava, Geoffrey Hinton, Alex Krizhevsky, Ilya Sutskever, and Ruslan Salakhutdinov.
\newblock Dropout: a simple way to prevent neural networks from overfitting.
\newblock \emph{The Journal of Machine Learning Research}, 15\penalty0 (1):\penalty0 1929--1958, 2014.

\bibitem[Wainwright \& Jordan(2008)Wainwright and Jordan]{wainwright2008graphical}
Martin~J Wainwright and Michael~I Jordan.
\newblock Graphical models, exponential families, and variational inference.
\newblock \emph{Foundations and Trends in Machine Learning}, 1\penalty0 (1--2):\penalty0 1--305, 2008.

\bibitem[Watson et~al.(2020)Watson, Lin, Klink, and Peters]{watson2020neural}
Joe Watson, Jihao~Andreas Lin, Pascal Klink, and Jan Peters.
\newblock Neural linear models with functional {G}aussian process priors.
\newblock In \emph{Third Symposium on Advances in Approximate Bayesian Inference}, 2020.

\bibitem[Watson et~al.(2021)Watson, Lin, Klink, Pajarinen, and Peters]{watson2021latent}
Joe Watson, Jihao~Andreas Lin, Pascal Klink, Joni Pajarinen, and Jan Peters.
\newblock Latent derivative {B}ayesian last layer networks.
\newblock In \emph{International Conference on Artificial Intelligence and Statistics}, 2021.

\bibitem[Welling \& Teh(2011)Welling and Teh]{welling2011bayesian}
Max Welling and Yee~W Teh.
\newblock Bayesian learning via stochastic gradient langevin dynamics.
\newblock In \emph{International Conference on Machine Learning}, 2011.

\bibitem[Wu et~al.(2019)Wu, Nowozin, Meeds, Turner, Hern{\'a}ndez-Lobato, and Gaunt]{wu2019deterministic}
Anqi Wu, Sebastian Nowozin, Edward Meeds, Richard~E Turner, Jos{\'e}~Miguel Hern{\'a}ndez-Lobato, and Alexander~L Gaunt.
\newblock Deterministic variational inference for robust {B}ayesian neural networks.
\newblock In \emph{International Conference on Learning Representations}, 2019.

\bibitem[Xiao et~al.(2017)Xiao, Rasul, and Vollgraf]{xiao2017fashion}
Han Xiao, Kashif Rasul, and Roland Vollgraf.
\newblock {Fashion-MNIST: A novel image dataset for benchmarking machine learning algorithms}.
\newblock \emph{arXiv:1708.07747}, 2017.

\bibitem[Yang et~al.(2024)Yang, Robeyns, Wang, and Aitchison]{yang2024bayesian}
Adam~X Yang, Maxime Robeyns, Xi~Wang, and Laurence Aitchison.
\newblock Bayesian low-rank adaptation for large language models.
\newblock In \emph{International Conference on Learning Representations}, 2024.

\bibitem[Zhang et~al.(2020)Zhang, Li, Zhang, Chen, and Wilson]{zhang2020cyclical}
Ruqi Zhang, Chunyuan Li, Jianyi Zhang, Changyou Chen, and Andrew~Gordon Wilson.
\newblock Cyclical stochastic gradient {MCMC} for {B}ayesian deep learning.
\newblock In \emph{International Conference on Learning Representations}, 2020.

\end{thebibliography}
\bibliographystyle{iclr2025}

\appendix
\section{How Deep Ensembles Secretly Perform Empirical Bayes}\label{app:extra_background}

The result that optimizing the ELBO in \autoref{eq:elbo} with flexible enough $\pi$ and $q$ results in $q^* = \pi^* = \pi^*(\, \cdot \mid \mathcal{D})$ with $q^*$ assigning probability $1$ to $\Theta^*$ might seem rather surprising at a first glance, since it is not often that a prior matches its posterior in Bayesian inference. This simple result can nonetheless be understood by simply inspecting \autoref{eq:elbo}: first, notice that $\pi$ appears only in the KL term, so that if the learnable prior $\pi$ is flexible enough, it must be the case that $\pi=q$ holds at optimality. It follows that $q$ must only maximize the first term in the ELBO, $\mathbb{E}_{\theta \sim q}[\log p(\mathcal{D} \mid \theta)]$; we can see by inspection that, when $q$ is flexible enough, $q^*$ must thus assign probability $1$ to $\Theta^*$. Additionally, when $q$ is flexible enough, it is well known from variational inference that maximizing the ELBO will result in the variational posterior matching the true posterior, i.e.\ $q = \pi(\, \cdot \mid \mathcal{D})$ should also hold at optimality. Combining these observations together, we get that $q^* = \pi^* = \pi^*(\, \cdot \mid \mathcal{D})$, where these distributions assign probability $1$ to $\Theta^*$. We refer the reader to \citet{loaiza-ganem2025deep} for a formal derivation of this result, along with a much more thorough discussion.

\section{Alternatives to Last Layer Empirical Bayes}\label{app:alternatives}

As mentioned in \autoref{sec:method}, we tried a few alternatives to LLEB which we now describe. First, we attempted to explicitly regularize the objective to encourage diversity. Since NFs admit density evaluation, it is straightforward to estimate the entropy $\mathbb{H}(q_\eta)$ of $q_\eta$. We thus attempted adding a regularizer which encourages maximizing entropy \citep{loaiza2017maximum}, resulting in the objective
\begin{equation}
    \mathbb{E}_{\theta_{QU} \sim q_\eta}\left[\log p(\mathcal{D} \mid \theta_{QU}, \theta_{NU})\right] + \lambda \mathbb{H}(q_\eta),
\end{equation}
which we maximized over $\theta_{NU}$ and $\eta$, where $\lambda > 0$ is a hyperparameter. We also tried the two-step solution we followed in LLEB, i.e.\ we first obtained $\theta_{QU}^*$ and $\theta_{NU}^*$ through maximum-likelihood, we discarded $\theta_{QU}^*$, and we then maximized $\mathbb{E}_{\theta_{QU} \sim q_\eta}[\log p(\mathcal{D} \mid \theta_{QU}, \theta_{NU}^*)] + \lambda \mathbb{H}(q_\eta)$ over $\eta$. Neither of these approaches improved upon LLEB as described in the main text.

After observing that encouraging higher entropy did not help, we hypothesized that maybe the NF was overly diverse to begin with and that it was not succeeding at placing most of its mass around $\Theta^*$. We thus tried using $\lambda < 0$ as a way of reducing diversity, but once again this did not improve upon LLEB.

Lastly, we also tried forgoing NFs entirely by replacing them with fully-connected architectures. We found that implementing this change resulted in $q^*$ collapsing onto a point mass, highlighting that the NFs used in LLEB indeed provide implicit regularization against this collapse. Note that since using fully-connected architectures loses density evaluation, we cannot regularize entropy to discourage this collapse.

These failed attempts are the reason why we used NFs and no entropy regularization in LLEB. Although we believe that LLEB can be improved upon by using different distributions which are flexible enough to concentrate mass on $\Theta^*$ while remaining as diverse as possible within $\Theta^*$, doing so is not trivial.

\section{Implementation Details}\label{app:details}

\paragraph{MNIST \& Fashion-MNIST} For these two datasets, we use a small convolutional network described in \autoref{tab:mnist_net}. For LLEB, we add to the weights of the last layer the output of a Neural Spline Flow \citep{durkan2019neural} implemented using the \texttt{nflow} library \citep{nflows} with parameters described in \autoref{tab:nsf_params}. For LLL, we use the implementation from \cite{daxberger2021laplace} on top of our network. For MCD, we keep the dropout layer active during evaluation. For LLEB, we use the end-to-end training objective with the reparameterization trick to maximize $\mathbb{E}_{\theta_{QU} \sim q_\eta}[\log p(\mathcal{D} \mid \theta_{QU}, \theta_{NU})]$, and use $10$ samples from $q_\eta$ per gradient step. 
At test time, for all three methods, we also sample $10$ times from $q^*$ and average the predictions to approximate the expectation in \autoref{eq:predictive}. The other training hyperparameters are given in \autoref{tab:mnist_params}.

\paragraph{CIFAR-10 \& SVHN} For both datasets, we use a ResNet18 \citep{he2016deep}, where we replace the last linear layer with two linear layers with a same hidden dimension of $50$, similarly to the network in \autoref{tab:mnist_net}. We use the hyperparameters described in \autoref{tab:cifar_params} to train the default network and ensemble. For LLEB, we use the two-step training procedure and use the frozen weights from the default network and train the flow for $100$ epochs and a  learning rate of $10^{-5}$. For MCD, since there are no dropout layers in ResNet18, we use the same frozen weights but use a new linear head preceded by a dropout layer, which we train until convergence ($10$ epochs, learning rate of $10^{-5}$).

\newpage

\begin{table}[t]
\centering
\caption{Network layers for MNIST and Fashion-MNIST.}
\begin{tabular}{l|l}
\textbf{Layer} & \textbf{Parameters} \\
\hline
Conv2d & input\_channels: 1, output\_channels: 10, kernel\_size: 5 \\
MaxPool2d & kernel\_size: 2, stride: 2 \\
ReLU &  \\
Conv2d & input\_channels: 10, output\_channels: 20, kernel\_size: 5 \\
Dropout2d &  \\
MaxPool2d & kernel\_size: 2, stride: 2 \\
ReLU &  \\
Flatten &  \\
Linear & input\_features: 320, output\_features: 50 \\
Linear & input\_features: 50, output\_features: 10 \\
\end{tabular}
\label{tab:mnist_net}
\end{table}

\begin{table}[t]
    \centering
    \caption{Parameters for the Neural Spline Flow.}
    \begin{tabular}{l|l}
        \textbf{Parameter} & \textbf{Value} \\
        \hline
        Base distribution & Gaussian \\
        Hidden features & 100 \\
        Number of coupling layers & 2 \\
        Number of residual blocks & 2\\
        Number of bins & 11 \\
        Tail bound & 10 \\
        Dropout probability & 0 \\
        Activations & ReLU
    \end{tabular}
    \label{tab:nsf_params}
\end{table}

\begin{table}[t]
    \centering
    \caption{Hyperparameters for training on MNIST and Fashion-MNIST.}
    \begin{tabular}{l|l}
        \textbf{Parameter} & \textbf{Value} \\
        \hline
        Epochs & 100 \\
        Optimizer & Adam \citep{kingma2015adam} \\
        Loss Function & Cross Entropy \\
        Learning Rate & $10^{-3}$ \\
        Weight Decay & $10^{-5}$ \\
        Gradient Clipping & 0.1 \\
        Batch Size & $10^{4}$
    \end{tabular}
    \label{tab:mnist_params}
\end{table}

\begin{table}[t]
    \centering
    \caption{Hyperparameters for training on CIFAR-10 and SVHN.}
    \begin{tabular}{l|l}
        \textbf{Parameter} & \textbf{Value} \\
        \hline
        Epochs & 100 \\
        Optimizer & Adam \\
        Loss Function & Cross Entropy \\
        Learning Rate & $5.10^{-4}$ \\
        Weight Decay & $10^{-5}$ \\
        Gradient Clipping & 0.1 \\
        Batch Size & $128$
    \end{tabular}
    \label{tab:cifar_params}
\end{table}

\end{document}